\def\year{2022}\relax
\documentclass[letterpaper]{article} 
\usepackage{aaai22}  
\usepackage{times}  
\usepackage{helvet}  
\usepackage{courier}  
\usepackage[hyphens]{url}  
\usepackage{graphicx} 
\usepackage{natbib}  
\usepackage{caption} 
\DeclareCaptionStyle{ruled}{labelfont=normalfont,labelsep=colon,strut=off} 
\usepackage{algorithm}
\usepackage{algorithmic}
\usepackage{amsmath}
\usepackage{amssymb}
\usepackage{newfloat}
\usepackage{listings}
\floatstyle{ruled}
\newfloat{listing}{tb}{lst}{}
\floatname{listing}{Listing}
\title{Dynamic Graph Learning-Neural Network for Multivariate Time Series Modeling}
\author{
    ZhuoLin Li,\textsuperscript{\rm 1}
    GaoWei Zhang,\textsuperscript{\rm 2}
    Jie Yu,\textsuperscript{\rm 1}
    LingYu Xu\textsuperscript{\rm 1,3}
}
\affiliations{
    \textsuperscript{\rm 1}School of Computer Engineering and Science, Shanghai University\\

    \textsuperscript{\rm 2}School of Artificial Intelligence, Beijing University of Posts and Telecommunications\\
    
    \textsuperscript{\rm 3}Shanghai Institute for Advanced Communication and Data Science, Shanghai University

}

\begin{document}

\maketitle

\begin{abstract}
Multivariate time series forecasting is a challenging task because the data involves a mixture of long- and short-term patterns, with dynamic spatio-temporal dependencies among variables. Existing graph neural networks (GNN) typically model multivariate relationships with a pre-defined spatial graph or learned fixed adjacency graph. It limits the application of GNN and fails to handle the above challenges. In this paper, we propose a novel framework, namely static- and dynamic-graph learning-neural network (SDGL). The model acquires static and dynamic graph matrices from data to model long- and short-term patterns respectively. Static matrix is developed to captures the fixed long-term association pattern via node embeddings, and we leverage graph regularity for controlling the quality of the learned static graph. To capture dynamic dependencies among variables, we propose dynamic graphs learning method to generate time-varying matrices based on changing node features and static node embeddings. And in the method, we integrate the learned static graph information as inductive bias to construct dynamic graphs and local spatio-temporal patterns better. Extensive experiments are conducted on two traffic datasets with extra structural information and four time series datasets, which show that our approach achieves state-of-the-art performance on almost all datasets.
\end{abstract}

\section{Introduction}
\noindent Multivariate time series forecasting has great applications in the fields of economics \cite{qin2017dual}, geographic \cite{liang2018geoman} and traffic \cite{yu2018spatio}. The challenge in the task is how to capture the interdependencies and dynamic evolutionary patterns among variables \cite{peng2020spatial}. Recently, spatio-temporal graph neural networks (GNN) have received increasing attention in modeling temporal data due to high capability in dealing with relational dependencies \cite{wu2020connecting}. 

Presently, most GNN methods are performed on data with a pre-defined graph structure \cite{guo2019attention,zheng2020gman,song2020spatial}. And now how to get the optimal graph structure becomes the latest research direction of GNN. Graph WaveNet \cite{wu2019graph} captures the hidden spatial dependencies by developing an adaptive dependencies matrix. IDGL \cite{chen2020iterative} finds the optimal graph structure by iteratively updating node embeddings and graph structures. The methods only work best with a well predefined graph. While in most multivariate time series data, the relationships between variables must be discovered from data, rather than provided as priori knowledge. MTGNN \cite{wu2020connecting} firstly study graph learning on multivariate time series without a priori structure.
\begin{figure}[t]
	\centering
	\includegraphics[width=0.9\columnwidth]{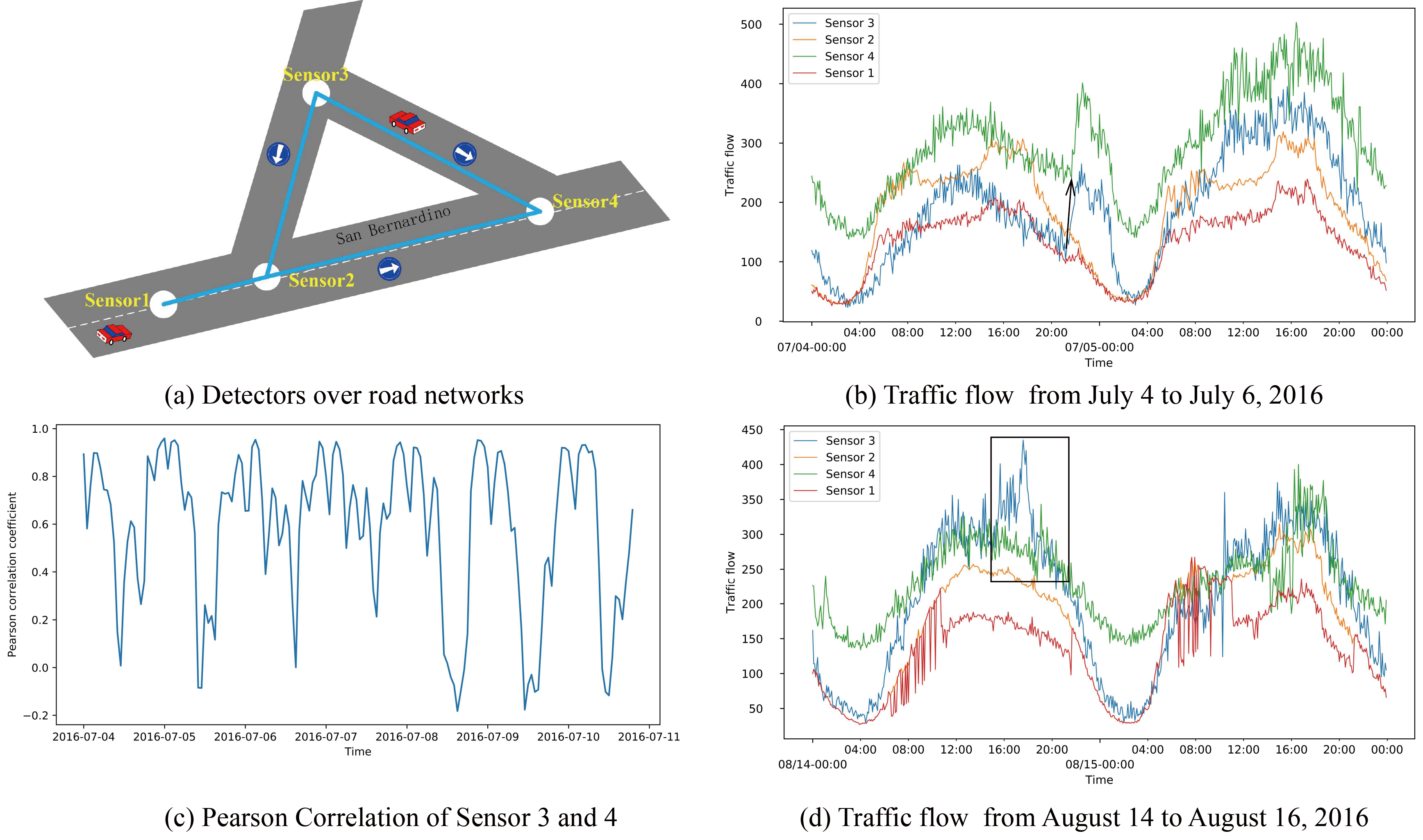} 
	\caption{Example of spatial-temporal dependencies in PeMSED8 dataset. (a) Practical location of the four sensors. (b) The traffic flow of sensor 4 is correlated to that of sensor 3, as shown by the black arrow. (c) Pearson Correlation coefficients for sensor 3 and 4. (d) Sudden events, as shown by the balck box}
	\label{fig1}
\end{figure}

The above methods seek a fixed graph structure, which cannot model spatial-temporal dependencies among variables explicitly and dynamically. Specifically, real-world applications often involve a mixture of long-term and short-term patterns. From Figure \ref{fig1}(a) and Figure \ref{fig1}(b), we can see that long-term correlation of data is dominated by the road network. Figure \ref{fig1}(c) indicates that there exist long-term patterns between sensors 3 and 4, but the short-term correlations between them are changing. Furthermore, there are sudden events, as shown in Figure \ref{fig1}(d). These factors lead to dynamic changes in spatial-temporal dependencies among variables, so the dependencies graph structure should be time-variant and reflect the interaction of patterns. In summary, GNN methods for multivariate time series modeling are facing the following challenges:
\begin{itemize}
\item \textbf{Challenge 1}: Interactions of long- and short-term patterns.
\item \textbf{Challenge 2}: Dynamic spatial-temporal dependencies among variables.
\end{itemize}

In this paper, we propose a novel approach to overcome these challenges. For challenge 1, our approach learns two types of graph matrices from data, namely static and dynamic graph matrices. Static graph is used to capture long-term patterns of data, in which we control the learning direction of graph through graph regularity. Dynamic graphs are taken to exact short-term patterns based on static graph, where static graph is constant and dynamic graphs are changing based on node-level data.

For challenge 2, we propose a dynamic graph learning approach to capture varying dependencies between variables. To capture the influence of long- and short-term patterns on inter-variate dependencies, we first utilize a gating mechanism to fuse both learned static patterns and dynamic input. On the basis, we design a multi-headed adjacency mechanism to efficiently extract the associations. To better model local changes, we add long-term patterns as inductive bias, which ensures that dynamic graphs fluctuate up and down around the long-term patterns. In summary, our main contributions are as follows:
\begin{itemize}
\item We propose a new graph learning framework to model the interaction of long- and short-term patterns of data, where static graph captures long-term pattern and dynamic graphs are used to model time-varying short-term patterns.
\item We design a dynamic graph learning method to efficiently dynamic capture spatial-temporal dependencies among variables, which mine the evolution of associations from data.
\item Experiments are conducted on two traffic datasets with predefined graph structures and four benchmark time series datasets. The results demonstrate that our method achieves state-of-the-art results on traffic datasets and most time series datasets without any prior knowledge.
\end{itemize}

\section{Related Works}
Time series forecasting has been studied for a long time. ARIMA\cite{li2012new} is a classical statistical method, while it cannot capture nonlinear relationships of data. GP \cite{frigola2015bayesian} make strong assumptions with respect to a stationary process and cannot scale well to multivariate time series data. Deep-learning-based methods can effectively capture non-linearity of data. Many of them employed convolutional neural networks (CNN) to exact dependencies between variables \cite{ma2019attention,sen2019think}. However, CNN cannot take full advantage of the dependencies between pairs of variables. Attention mechanisms are used in time series modeling \cite{huang2019dsanet} due to the ability to learn feature weights adaptively. However, it models temporal and spatial correlations separately, cannot fully exploit dependencies between variables.

Graph neural networks(GNN) have enjoyed success in handling dependencies among entities, where most methods assume that a well-defined graph already exists \cite{kipf2016semi,velivckovic2017graph}. Now, researchers work to discover the optimal graph structure from data to improve the performance of GNN \cite{wu2019graph}. AM-GCN \cite{wang2020gcn} compensates for the limitations of predefined graph by producing a new feature graph. GCNN \cite{Diao2019dynamic} proposed a dynamic matrix estimator to track the spatial dependencies of data. However, the method requires pre-training of the Tensor Decomposition Layer. STFGNN \cite{li2021spatial} proposed a "temporal graph" approach to compensate for existing correlations. The method requires prior construction of the "temporal graph" and an additional operation to obtain spatial-temporal dependencies of data. The performance of the above methods is heavily relying on the quality of prior graph.

\section{Preliminaries}
Let $x^t\in \mathbb{R}^N $ denote the value of a multivariate variable of dimension $N$ at time step $t$, where $x^t[i]\in \mathbb{R}$ denotes the value of $i^{th}$ variable at time step $t$. Given a sequence of historical $H$ time steps of observations on a multivariate variable, $X={\{x^{t_1},x^{t_2},...,x^{t_h}\}} = \{ x_1, x_2,..., x_N\} \in \mathbb{R}^{N \times h}$, our goal is to predict the $L$-step-away value of $Y={\{x^{t_{h+L}}\}}$, or a sequence of future values $Y={\{x^{t_{h+1}},x^{t_{h+2}},...,x^{t_{h+L}}\}}$. We aim to build a mapping $f(.)$ from $X$ to $Y$, ie., $Y=f(X)$.

\begin{figure}[t]
	\centering
	\includegraphics[width=0.9\columnwidth]{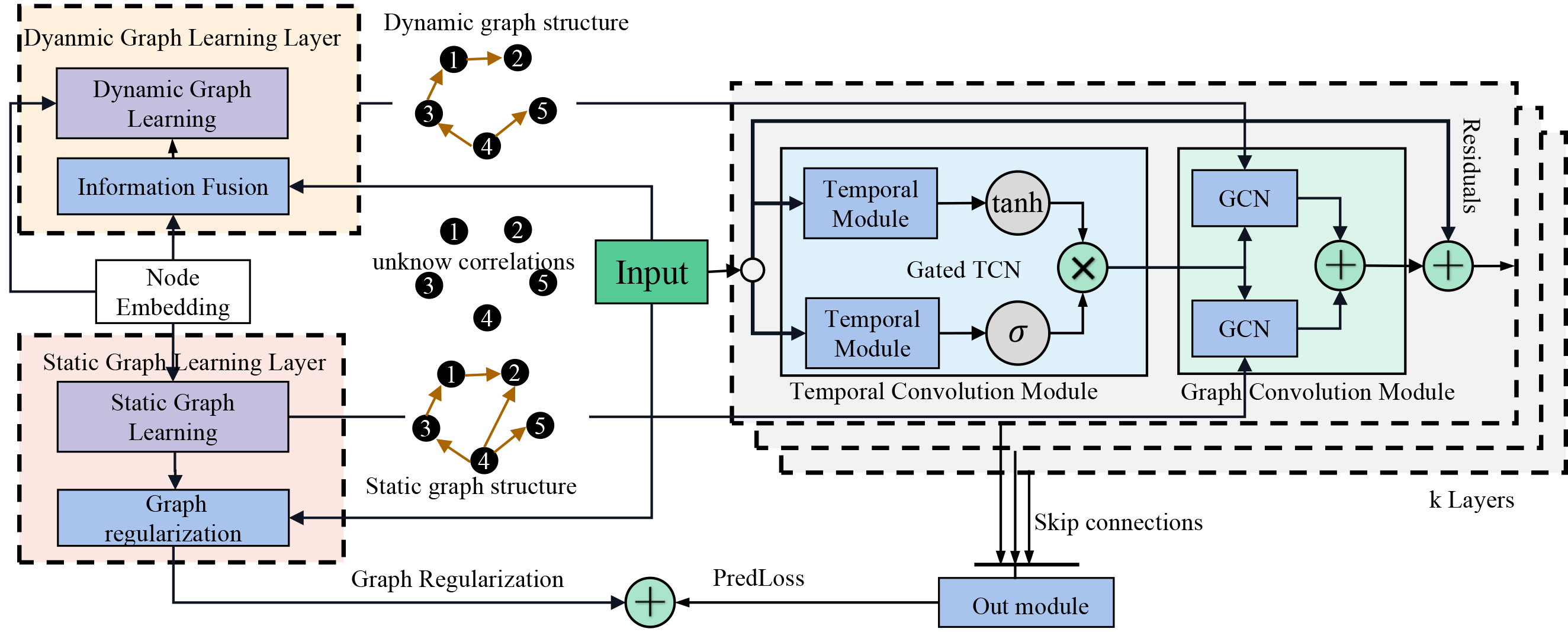} 
	\caption{The framework of SDGL. The model consists of static-, dynamic- graph learning layer, temporal convolution (TCN) and graph convolution module (GCN).  Input is first passed through graph learning layers to obtain the graph structure. Then input and graphs go through the k-layer TCN and GCN for feature transformation. Finally, the graph regularization and predicted loss are the final loss.}
	\label{fig2}
\end{figure}
\section{Framework of SDGL}
We first elaborate on the general framework of our model. As illustrated in Figure \ref{fig2}, SDGL on the highest level consists of static and dynamic graph learning layers, temporal convolution (TCN), graph convolution modules (GCN), and an output module. To discover the hidden dependencies between variables, two graph learning layers generate two kinds of adjacency matrices, i.e., static and dynamic matrices. TCN adopts a gated structure, which consists of two parallel temporal modules, to extract the temporal dependencies. In GCN, we use two separate modules to aggregate information based on learned static and dynamic matrices. Figure \ref{fig2} shows how the each module collaborates with each other. In more detail, the core components of our model are illustrated in the following.

\subsection{Static Graph Learning Layer}
Follow previous work \cite{bai2020adaptive,wu2020connecting}, we employ node embeddings to capture fixed associations in the data, independent of the dynamic node-level input. The difference is that we add graph regularity to control the learning direction during training, which makes the learned graphs have better convergence and interpretability.

\textbf{Static Graph Learning}: In static graph learning, we first randomly initialize a learnable node embeddings dictionary $M_s \in \mathbb{R}^{N \times d}$, where $d$ is the dimension of node embeddings. Then we infer the dependencies between each pair of nodes by Eq.(\ref{Eq:1}).
\begin{equation}
	\hat{A}=SoftMax(ReLU(M_s\cdot{M_s}^\top))
	\label{Eq:1}
\end{equation}

We use ReLU activation function to eliminate weak connections. Softmax function is used to normalize the adaptive matrix.

\textbf{Graph Regularization}: It is important to control the smoothness, connectivity and sparsity of learned graph, so we add graph regularization \cite{jiang2019semi} to control the quality of resulting graph. We first apply regularization to dynamic datasets to control graph learning direction. And instead of applying regularization to all inputs at once, we apply it to the data in the gradient update section each time. Given input $X=(x_1,x_2,...,x_N) \in \mathbb{R}^{N \times h}$ and $\hat{A}$, the regularization formula is as follows:
\begin{equation}
	L_{GL} = \sum_{i,j=1}^{N}{\parallel x_i-x_j \parallel}_2^2 \hat{A} + \gamma\parallel\hat{A}\parallel_F^2
	\label{Eq:2}
\end{equation}
It can be seen that minimizing the first term forces adjacent nodes to have similar features, thus enforcing smoothness of the graph signal on the graph associated with $\hat{A}_{ij}$. The first term in Eq.(\ref{Eq:2}) only requires time window length $H$ data $X$, but during the training, all data in training set affect $\hat{A}$. Eventually, the $\hat{A}_{ij}$ is given a large value if only $x_i$ and $x_j$ are similar throughout the time range covered by the training set, which ensures that static graph captures long-term dependencies among variables. The second term, where $\parallel\hat{A}\parallel_F^2$ denotes the Frobenius norm of the matrix, controls sparsity by penalizing large degrees due to the first term in Eq.\ref{Eq:1}. 

\subsection{Dynamic Graph Learning Layer}

In multivariate time series data, the dependencies between variables are dynamically changing, which are affected by both fixed graph structure and dynamic inputs, e.g., traffic congestion upstream will gradually affect traffic downstream over time. Different current methods that only exploit graph structure \cite{wu2020connecting} or dynamic inputs \cite{li2021spatial} to get a fixed matrix, we design a dynamic graph learning layer (DGLL) which accounts for both information and dynamically capture spatial-temporal associations between variables. We first fuse the graph structure and input information using designed Information Fusion module. Then we propose Dynamic Graph learning Module to dynamically generate matrices based on fusion results. Due to short-term patterns tend to be locally changing, we add learned fixed patterns as induction bias.

\subsubsection{Information Fusion module}
To adaptively fuse the fix graph and dynamic inputs, we apply a gated mechanism \cite{cholearning}. Given the node embeddings $M_s \in \mathbb{R}^{N \times d}$ and input $X$. We firstly use a linear layer to transform $X$ to $X_T$ which has the same dimension as $M_s$. Then information fusion formula is as follows:
\begin{align}
	r_T=\delta(W_{r}\cdot M_s+U_r \cdot X_T ) \notag\\
	z_T=\delta(W_z \cdot M_s+U_z \cdot X_T)\notag\\
	\hat{h}_T=tanh(W_h \cdot X_T + r_T{(U_h\cdot M_s)})\notag\\
	h_T = (1-z_T) \cdot M_s + z_T \cdot \hat{h_T}
	\label{Eq:3}
\end{align}
Where $W$ and $U$ in Eq.(\ref{Eq:3}) are weight matrices need be updated. Then we get result $h_T \in \mathbb{R}^{N \times d}$.
\begin{figure}[t]
	\centering
	\includegraphics[width=0.8\columnwidth]{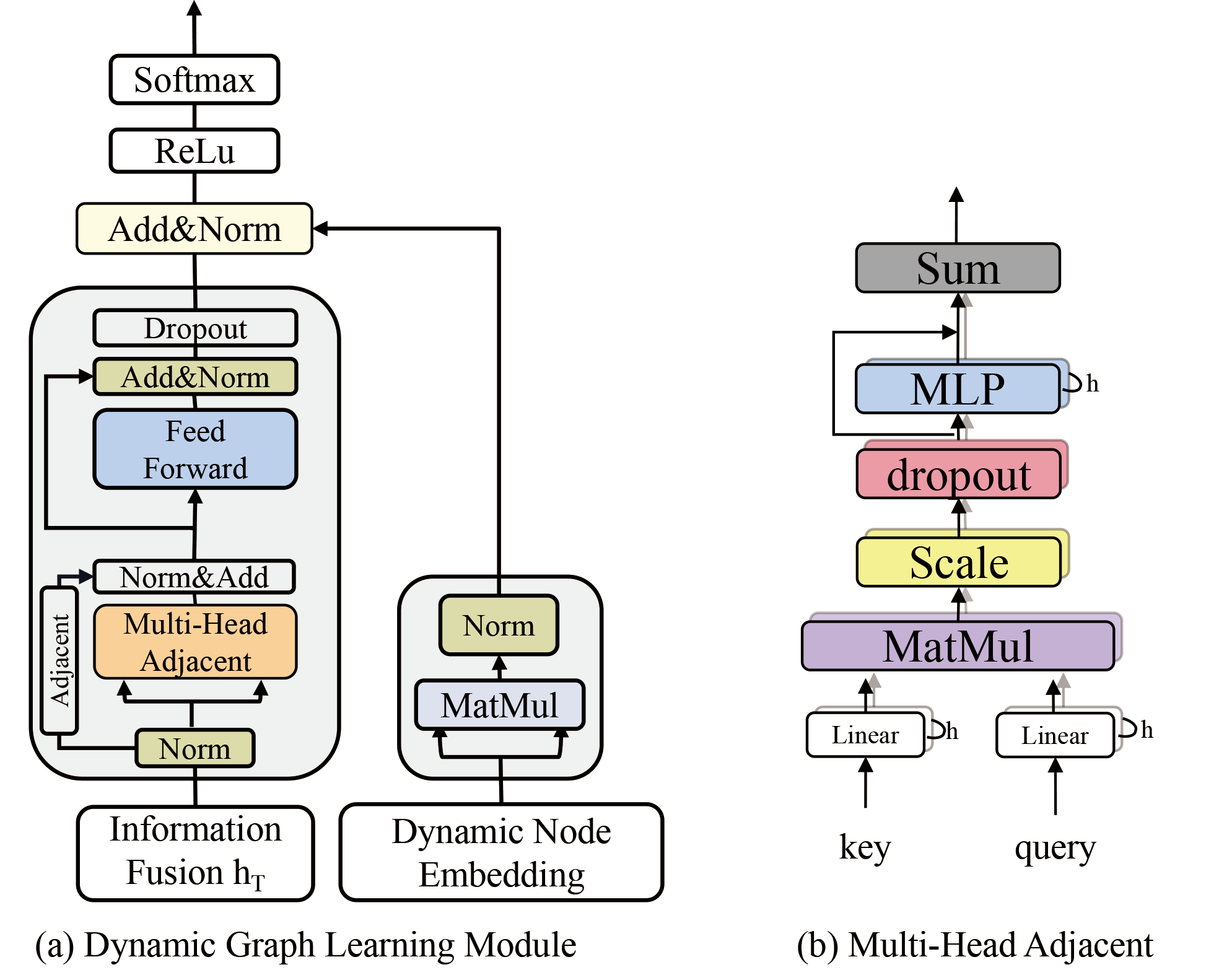} 
	\caption{The framwork of Dynamic Graph Learning Module}
	\label{fig3}
\end{figure}
\subsubsection{Dynamic Graph Learning Module}
To effectively capture dynamic short-term dependencies between variables, we design dynamic graph learning module (DGLM) that directly produces associations matrices according to the fusion result. As shown in Figure \ref{fig3}(a), DGLM consists of two main components: dynamic matrix construction on the left and inductive bias on the right. The framework of DGLM is inspired by self-attention \cite{vaswani2017attention}, but differs from self-attention, our goal is to construct the association matrices between nodes. And we add fixed structure information as inductive bias instead of self-attention without any prior assumptions, which can model spatial-temporal dependencies faithfully.

\textbf{Dynamic relationship construction:} The input $h_T$ is first passed through a layer normalization in order to train the model easier\cite{xiong2020layer}. We adopt a multi-head way to construct a matrix to capture correlations among variables from multiple perspectives, as shown in Figure \ref{fig3}(b). We project $Q$, $K$ with $W_q$ and $W_k$ linear projections onto the $d_k$, $d_k$ dimensions $head$ times, where $K$ and $Q \in \mathbb{R}^{N \times head \times d_k}$. Then we perform matrix computation in parallel on each projected version of $Q$, $K$ and produce $head$ matrices, where we adopt scaled dot-product to compute the correlation between $v_i$ and $v_j$.
\begin{equation}
	Adj_{{v_i},{v_j}}^{head_i}=dropout(\frac{Q_{head_i}{K_{head_i}^\top}}{\sqrt{d_k}})
	\label{Eq:4}
\end{equation}
In Eq.(\ref{Eq:4}), we added dropout to improve generalization performance of model, as shown in Figure \ref{fig3}(b). Then we projected by a linear layer to get $head_i= W_hAdj_{v_i,v_j}^{head_i}$. Finally the multi-head matrices are summed to get $R_T \in \mathbb{R}^{N \times N}$,
\begin{equation}
		R_T = Sum(head_1,...,head_n),
		\label{Eq:5}
\end{equation}

To avoid that output $R_T$ converges to a rank-1 matrix \cite{dong2021attention}, we add skip connection and multi-layer perceptrons (MLPs). 
We do dot-product using $E_r \in \mathbb{R}^{N \times d_k}$, where $E_r=W_rh_T$, as residual connection as in Eq.(\ref{Eq:6}), where $LN$ denotes LayerNorm. Then we exploit MLPs to feature projection, as shown in Eq.(\ref{Eq:7}). 
\begin{equation}
	S_T = LN(R_T)+LN(E_r \cdot {E_r}^\top)\\
	\label{Eq:6}
\end{equation}
\begin{equation}
	\hat{S}_T = max(0, S_TW_{1}+b_1)W_{2}+b_2
	\label{Eq:7}
\end{equation}

\textbf{Inductive bias:} Spatial-temporal dependencies in short-term patterns tends to locally change, e.g. peak period traffic patterns tend to be variable but related to road structure. To reflect such trends, we add induction bias, as shown in Eq.(\ref{Eq:8}).
\begin{equation}
	A_d^T = Softmax(Relu(LN(M_d \cdot {M_d}^\top) + \hat{S}_T))
	\label{Eq:8}
\end{equation}
where $M_d \in \mathbb{R}^{N \times d_k} $ is dynamic node embeddings, from Eq.(\ref{Eq:9}). The dot result of $M_d$ provides fixed structure information for dynamic matrices. Details of $M_d$ are shown below.

\textbf{Momentum update:} Our graph structure learning method relies on node embeddings, so the quality of it is crucial. To learn static and dynamic matrices jointly and avoid the opposite effect of two types of graph structures on the same node embeddings, we use two independent node embeddings. Also to ensure the association of dynamic node embeddings with static node embeddings, we introduced momentum update \cite{he2020momentum}. The update equation is shown below.
\begin{equation}
	M_d \leftarrow pM_d + (1-p)M_s
	\label{Eq:9}
\end{equation}
Here $p \in [0,1) $ is a momentum coefficient. Only the parameters $M_s$ are updated by back-propagation. The process of dynamic graph learning is shown in Algorithm \ref{alg:1}.
\begin{algorithm}[tb]
	\caption{Dynamic Graph Learning}
	\label{alg:1}
	\textbf{Input}: $X \in \mathbb{R}^{N \times h}$\\
	\textbf{Parameter}: node embeddings $M_s \in \mathbb{R}^{N \times d_k}, M_d \in \mathbb{R}^{N \times d_k}$\\
	\textbf{Output}: dynamic graph $A_d^T$
	\begin{algorithmic}[1] 
		\STATE $h_T \leftarrow X$ and $M_s$ using Eq.(\ref{Eq:3}) \{get fusion result of $T$\}
		\STATE $Q$, $K$, $E_r \leftarrow h_T$ by linear layer 
		\STATE $R_T \leftarrow \{Q, K\}$ using Eq.(\ref{Eq:4}) and Eq.(\ref{Eq:5})
		\STATE $\hat{S}_T \leftarrow \{R_T, E_r\}$ using Eq.(\ref{Eq:6}) and Eq.(\ref{Eq:7})
		\STATE $A_d^T \leftarrow \{\hat{S_T}, M_d\}$ using Eq.(\ref{Eq:8}) \{add inductive bias\}
		\STATE \textbf{return} $A_d^T$
	\end{algorithmic}
\end{algorithm}
\subsection{Temporal Convolution Module}
In temporal convolution module, we take a cnn-based method for the benefits of parallel computing, stable gradients. As shown in Figure \ref{fig2}, we employ gating mechanism \cite{dauphin2017language}, dilated convolutions \cite{vanwavenet} and inception \cite{szegedy2015going}. We let the dilation factor for each layer increase exponentially at a rate of $q$ ($q>1$). Suppose the initial factor is 1 and $m$ 1D convolution layers of kernel size $c$, the receptive field of $k$ layer dilated convolutional network is:
\begin{equation}
	R = 1+(c-1)(q^k-1)/(q-1)
	\label{Eq:10}
\end{equation}
In the work, inception layer consisting of four filter sizes, viz. $1 \times 2$, $1 \times 3$, $1 \times 6$, $1 \times 7$, which the largest kernel size is used for the calculation the receptive field. Given a 1D sequence input $x \in \mathbb{R}^T$ and filters consisting of $f_{1 \times 2} \in \mathbb{R}^2$,  $f_{1 \times 3} \in \mathbb{R}^3$, $f_{1 \times 6} \in \mathbb{R}^6$, $f_{1 \times 7} \in \mathbb{R}^7$, the dilated inception layer takes the form Eq.(\ref{Eq:11}).
\begin{equation}
	H = concat(H * f_{1 \times 2}, H * f_{1 \times 3}, H * f_{1 \times 6}, H * f_{1 \times 7})
	\label{Eq:11}
\end{equation}
Where the outputs of the four filters are truncated to the same length according to the largest filter and concatenated across the channel dimension. And the dilated convolution denoted by $H * f_{1 \times r}$ is defined as $	H * f_{1 \times r}(t) = \sum_{s=0}^{r-1}f_{1 \times r}(s)H(t-d \times s)$, where d is the dilation factor.

\subsection{Graph Convolution Module}
The graph convolution module aims to fuse node’s with its neighbors’ features to obtain new node representation. Li et al. modeled diffusion process of graph signals with $S$ finite steps \cite{li2018diffusion}. Let $A \in \mathbb{R}^{N \times N}$ denote the adjacency matrix, $X \in \mathbb{R}^{N \times D}$ denote the input, $Z \in \mathbb{R}^{N \times M}$ denotes result and $W_s \in \mathbb{R}^{D \times M}$ denote model parameters. The form can be generalize into $Z=\sum_{s=0}^{s}P^sXW_s$, where $P^s\in \mathbb{R}^{N \times N}$ represents the power series of transition matrix. In the case of an graph, $P = A/rowsum(A)$. 

We decouple the information propagation and representation transformation operations \cite{liu2020towards} and exploit the $W_s$ as the information selection layer, with the following equation:
\begin{equation}
	Z = W_sconcat(P^1X,...P^sX, X)
	\label{Eq:12}
\end{equation}
where $W_s$ implemented with $1 \times 1$ convolutional, with the input channel $c(s+1)$ and output channel $c$. In the extreme case, i.e., where there is no dependence among variables, aggregating information just adds noise to each node. So, we use $W_s$ to select the significant information generated at each transition matrix. In above case, Eq.(\ref{Eq:12}) still preserves nodes' own information by adjusting $W_s$ to 0 for $P^1X,...P^sX$.

To model the interaction of short-term and long-term patterns, we adopt two graph convolution layers to capture node information on static and dynamic graphs separately, that is, replacing A with learned $\hat{A}$ and $A_d^T$. The final output is shown in Eq.(\ref{Eq:13}).
\begin{equation}
	Z_f = Z_{static} + Z_{dynamic}
	\label{Eq:13}
\end{equation}
\subsection{Joint Learning with A Hybrid Loss}
In contrast to previous work that optimizes the adjacency matrix based on task-related prediction loss, we jointly learn the graph and GNN parameters by minimizing a hybrid loss that combines prediction and graph regularization loss.

The full process of SDGL is shown in Algorithm \ref{Eq:2}. As we can see, our framework only needs to initialize node embeddings $M_s$. Then we construct static graph based on $M_s$ (Eqs. (1)), and dynamic graphs are generated using Algorithm \ref{alg:1}. The input $X$ and learned graphs are subjected to feature extraction and aggregation by TCN and GCN (Eq.(\ref{Eq:11}) and Eq.(\ref{Eq:13})). Finally, we update parameters of model according to hybrid loss $L_{loss}$, while updating the $M_d$ using momentum update (Eq. (\ref{Eq:9})).
\begin{algorithm}[tb]
	\caption{General Framework of SDGL}
	\label{alg:2}
	\textbf{Input}: The dataset $O$ \\
	\textbf{Parameter}: $M_s$, $M_d$, $H$, $L$, $iter$, $K$, $B$, $epoch$\\
	\textbf{Output}: prediction value $\hat{Y}$
	\begin{algorithmic}[1] 
		\STATE init node embeddings and $M_d \leftarrow M_s$
		\FOR{$r$ in $1: epoch$}
		\FOR{$i$ in $1:iter$}
		\STATE Sample $X \in R^{B\times N\times H}$, $Y \in R^{B\times N\times L}$ from $O$
		\STATE calculate received filed using Eq.(\ref{Eq:10})
		\STATE $\hat{A} \leftarrow \{M_s\}$ using Eq.(\ref{Eq:1})
		\STATE $L_{GL} \leftarrow \{\hat{A}, X\}$ using Eq.(\ref{Eq:2})
		\STATE $A_d^T \leftarrow\{M_s, M_d, X\}$ using Algorithm.(\ref{alg:1})
		\FOR {$j$ in $1:K$}
		\STATE $H^j \leftarrow \{X\}$ using Eq.(\ref{Eq:11})
		\STATE $Z_f^j \leftarrow \{H^j, \hat{A}, A_d^T\}$ using Eq.(\ref{Eq:13})
		\STATE $X \leftarrow Z_f^j$ and $output += H^j$
		\ENDFOR
		\STATE $\hat{Y} \leftarrow \{ output\}$ using output module
		\STATE $L_{loss} \leftarrow \lambda L_{GL} + L1loss(\hat{Y}, Y)$
		\STATE Back-propagate $L_{loss}$ update model paramters and $M_d$ using Eq.(\ref{Eq:9}), $i \leftarrow i+1$
		\ENDFOR
		\STATE $r \leftarrow r+1$
		\ENDFOR
	\end{algorithmic}
\end{algorithm}

\section{Experiments}
To evaluate the performance of SDGL, we conduct experiments on two tasks single-step and multi-step forecasting. First, to illustrate how well SDGL performs, we evaluate it on two public traffic datasets \cite{bai2020adaptive}, compared to other spatio-temporal graph neural networks, where the aim is to predict multiple future steps. Then, we compare the SDGL with other multivariate time series models on four benchmark datasets, where the aim is to predict a single future step \cite{wu2020connecting}.

\subsection{Dataset and Experimental Setting}
\begin{table}[t]
	\centering
	\resizebox{.95\columnwidth}{!}{
	\begin{tabular}{ccccc}
		\hline
		Datasets&Samples&Nodes&Input\_len&Pred\_len\\
		\hline
		PeMSD4&16992&307&12&12\\
		PeMSD8&17856&170&12&12\\
		\hline
		Traffic&17544&862&168&1\\
		Solar-energy&52560&137&168&1\\
		Electricity&26304&321&168&1\\
		Exchange-rate&7588&8&168&1\\
		\hline
	\end{tabular}
	}
	\caption{Dataset statistics}
	\label{table1}
\end{table}
In Table 1, we summarize the statistics of benchmark datasets. In Multi-step prediction experiment, following the method \cite{bai2020adaptive}, we use three evaluation metrics, i.e., MAE, MAPE and RMSE. In Single-step prediction experiment, we use Root Relative Squared Error (RSE), and Coefficient Correlation (CORR) to measure the performance of predictive models \cite{wu2020connecting}. The specific calculation formula is detailed in Appendix A.2. For RMSE, MAE, MAPE and RRSE, lower the values are better. For CORR, higher the values are better.

\subsection{Baseline Methods for Comparision}
\subsubsection{Multi-step forecasting}
\begin{itemize}
	\item HA: Historical Average, which uses the average of previous seasons as the prediction.
	\item VAR \cite{zivot2006vector}: Vector Auto-Regression:a model that captures correlations among traffic series.
	\item DSANet \cite{huang2019dsanet}: A prediction model using CNN networks and self-attention mechanism.
	\item DCRNN \cite{li2018diffusion}: A diffusion convolutional recurrent neural network containing graph convolution.
	\item STGCN \cite{yu2018spatio}: A spatial-temporal GNN.
	\item ASTGCN \cite{guo2019attention}: Attention-based spatio-temporal GNN, which integrates spatial and temporal attention mechanisms.
	\item STSGCN \cite{song2020spatial}: Spatial-temporal GNN that captures correlations by stacking multiple localized GCN layers with adjacent matrix over the time axis.
	\item AGCNR \cite{bai2020adaptive}: Adaptive Graph Convolutional Recurrent Network that capture association of data and node-specific patterns through node embeddings.
\end{itemize}
\subsubsection{Single-step forecasting}
\begin{itemize}
	\item AR: Auto-regressive model, a statistical model.
	\item VAR-MLP \cite{zhang2003time}: A hybrid model of the multilayer perception  and auto-regressive model.
	\item GP \cite{frigola2015bayesian} A Gaussian Process time series model.
	\item RNN-GRU: A recurrent neural network with fully connected GRU hidden units.
	\item LSTNet \cite{lai2018modeling}: Long- and Short-term Time-series network, which uses the CNN and Recurrent Neural Network to extract short-term patterns and to discover long-term patterns for time series trends.
	\item TPA-LSTM \cite{shih2019temporal}: Temporal pattern attention network that capture patterns across time steps.
	\item MTGNN \cite{wu2020connecting}: The first study on multivariate time series from a graph-based perspective with GNN.
\end{itemize}	

\subsection{Main Results}
\subsubsection{Multi-step forecasting}
\begin{table*}[t]
	\centering
	\resizebox{.95\textwidth}{!}{
		\begin{tabular}{cc*{10}{c}}
			\hline
			Dataset & Metric& HA& VAR& GRU-ED& DSANet& DCRNN& STGCN& ASTGCN& STSGCN& AGCNR& SDGL\\ 
			\hline
			& MAE& 38.03&24.54&23.68&22.79&21.22&21.16&22.93&21.19&\underline{19.83}&\textbf{18.65}\\
			PeMSD4& MAPE(\%)&27.88&17.24&16.44&16.03&14.17&13.83&16.56&13.9&\underline{12.97}&\textbf{12.38}\\
			& RMSE& 59.24&38.61&39.27&35.77&33.44&34.89&35.22&33.65&\underline{32.26}&\textbf{31.30}\\
			\hline
			\hline
			& MAE& 34.86&19.19&22&17.14&16.82&17.5&18.25&17.13&\underline{15.95}&\textbf{14.93}\\
			PeMSD8& MAPE(\%)& 24.07&13.1&13.33&11.32&10.92&11.29&11.64&10.96&\underline{10.09}&\textbf{9.61}\\
			&RMSE&52.04&29.81&36.23&26.96&26.36&27.09&28.06&26.86&\underline{25.22}& \textbf{24.13}\\
			\hline
		\end{tabular}
	}
	\caption{Baseline comparison under multi-step forecasting for spatial-temporal graph neural networks.}
	\label{table2}
\end{table*}
Table \ref{table2} and Table \ref{table3} provide the main experimental results of SDGL. We observe that our model achieves state-of-the-art results on most of tasks, with on-par performance comparable to the state-of-the-art in the remaining tasks. In the following, we discuss experimental results of multi-step and single-step forecasting respectively.

\subsection{Multi-step forecasting}
Table \ref{table2} presents the overall prediction performances, which are the averaged MAE, RMSE and MAPE over 12 prediction horizons, of our SDGL and night representative comparison methods. We can observe that: 1) GCN-based methods outperform baselines and self-attention-based DSANet, demonstrating the effectiveness of GCN in traffic time series forecasting. 2) The performance of the graph-learned method performs better than that of using predefined graphs. AGCNR significantly improves the performance of GCN-based method based on learned dependencies. 3) Our method further improves learning-graph based methods with a significant margin. SDGL brings more than 6\% relative improvements to the existing best results in MAE for PeMSD4 and PeMSD8 datasets. 

\subsection{Single-step forecasting}
In this experiment, we compare SDGL with other multivariate time series models. Table \ref{table3} shows the experimental results for the single-step forecasting task. 1) The performance of the methods that model the relationships between variables show higher performance, i.e., TPA-LSTM, demonstrating the importance of modeling the relationships. 2) GNN-based methods outperform other baseline methods, i.e., MTGNN, indicating that GNN has greater ability to model dependencies between variables. 3) Our approach further improves the performance of GNN-based method, which demonstrates the effectiveness of SDGL. On Solar-Energy and Electricity dataset, the improvement of SDGL in terms of RSE is significant, which SDGL lowers down RSE by more than 4\% over the horizons of 3, 6, 12, 24 on the Solar-Energy data. SDGL slightly outperforms MTGNN in traffic data, where MTGNN improves traffic forecasting performance significantly. We conjecture the reason is that the target of the forecast is road occupancy, which the values changes with relatively stable patterns. SDGL improves on CORR metrics for exchange rate data only, but outperforms the graph neural network MTGNN in all metrics.
\begin{table*}[t]
	\centering
	\resizebox{\textwidth}{!}{
		\begin{tabular}{cc|*{4}{c}|*{4}{c}|*{4}{c}|*{4}{c}} 
			\hline
			\multicolumn{2}{c}{Dataset}&\multicolumn{4}{c}{Solar\_Energy}& \multicolumn{4}{c}{Traffic}&\multicolumn{4}{c}{Electricity}&\multicolumn{4}{c}{Exchange\_Rate}\\
			\hline
			&&\multicolumn{4}{c}{Horizon}&\multicolumn{4}{c}{Horizon}& \multicolumn{4}{c}{Horizon}&\multicolumn{4}{c}{Horizon}\\
			\hline
			Methods& Metrics&3&6&12&24&3&6&12&24&3&6&12&24&3&6&12&24\\
			\hline
			AR& RSE&0.2435&0.3790&0.5911&0.8699&0.5991&0.6218&0.6252&0.63&0.0995&0.1035&0.1050&0.1054&0.0228&0.0279&0.0353&0.0445\\
			& CORR&0.9710&0.9263&0.8107&0.5314&0.7752&0.7568&0.7544&0.7519&0.8845&0.8632&0.8591&0.8595&0.9734&0.9656&0.9526&0.9357\\
			\hline
			VARMLP& RSE&0.1922&0.2679 &0.4244 &0.6841&0.5582&0.6579&0.6023&0.6146&0.1393&0.1620&0.1557&0.1274&0.0265&0.0394&0.0407&0.0578\\
			&CORR&0.9829&0.9655&0.9058&0.7149&0.8245&0.7695&0.7929&0.7891&0.8708 &0.8389 &0.8192 &0.8679&0.8609 &0.8725 &0.8280&0.7675\\
			\hline
			GP&RSE&0.2259&0.3286&0.5200&0.7973&0.6082&0.6772&0.6406&0.5995&0.1500 &0.1907 &0.1621 &0.1273 &0.0239 &0.0272 &0.0394 &0.0580\\
			&CORR&0.9751&0.9448&0.8518&0.5971&0.7831&0.7406&0.7671&0.7909&0.8670&0.8334&0.8394&0.8818 &0.8713 &0.8193 &0.8484&0.8278\\
			\hline
			RNN-GRU&RSE&0.1932&0.2628&0.4163&0.4852&0.5358&0.5522&0.5562&0.5633&0.1102&0.1144&0.1183&0.1295&0.0192&0.0264&0.0408&0.0626\\
			&CORR&0.9823&0.9675&0.9150&0.8823&0.8511&0.8405&0.8345&0.8300&0.8597&0.8623&0.8472&0.8651&0.9786&\underline{0.9712}&0.9531&0.9223\\
			\hline
			LSNet-skip&RSE&0.1843&0.2559&0.3254&0.4643&0.4777&0.4893&0.4950&0.4973&0.0864&0.0931&0.1007&0.1007&0.0226&0.0280&0.0356&0.0449\\
			&CORR&0.9843&0.9690&0.9467&0.8870&0.8721&0.8690&0.8614&0.8588&0.9283&0.9135&0.9077&0.9119&0.9735&0.9658&0.9511&0.9354\\
			\hline
			TPA-LSTM&RSE&0.1803&\underline{0.2347}&0.3234&0.4389&0.4487&\underline{0.4658}&0.4641&0.4765&0.0823&0.0916&0.0964&0.1006&\textbf{0.0174}&\textbf{0.0241}&\textbf{0.0341}&\textbf{0.0444}\\
			&CORR&0.9850&\underline{0.9742}&0.9487&\underline{0.9081}&0.8812&\underline{0.8717}&0.8717&0.8629&0.9439&0.9337&0.9250&0.9133&\underline{0.9790}&0.9709&\underline{0.9564}&\underline{0.9381}\\
			\hline
			MTGNN&RSE&\underline{0.1778}&0.2348&\underline{0.3109}&\underline{0.4270}&\underline{0.4162}&0.4754&\textbf{0.4461}&\textbf{0.4535}&\underline{0.0745}&\underline{0.0878}&\underline{0.0916}&\underline{0.0953}&0.0194&0.0259&0.0349&0.0456\\
			&CORR&\underline{0.9852}&0.9726&0.9509&0.9031&0.8963&0.8667&\textbf{0.8794}&\textbf{0.8810}&\underline{0.9474}&\underline{0.9316}&\underline{0.9278}&\underline{0.9234}&0.9786&0.9708&0.9551&0.9372\\
			\hline
			\hline
			SDGL&RSE&\textbf{0.1699}&\textbf{0.2222}&\textbf{0.2924}&\textbf{0.4047}&\textbf{0.4142}&\textbf{0.4475}&0.4584&0.4571&\textbf{0.0698}&\textbf{0.0805}&\textbf{0.0889}&\textbf{0.0935}&0.0180&0.0249&0.0342&0.0455\\
			&CORR&\textbf{0.9866}&\textbf{0.9762}&\textbf{0.9565}&\textbf{0.9119}&\textbf{0.9010}&\textbf{0.8825}&0.8760&0.8766&\textbf{0.9534}&\textbf{0.9445}&\textbf{0.9351}&\textbf{0.9301}&\textbf{0.9808}&\textbf{0.9730}&\textbf{0.9583}&\textbf{0.9402}\\
			\hline
		\end{tabular}
	}
	\caption{Baseline comparison under single-step forecasting for multivariate time series methods.}
	\label{table3}
\end{table*}

\subsection{Ablation Study}
We conduct an ablation study on the PeMSD and PeMSD8 datasets to validate the effectiveness of key components that contribute to the improved outcomes of our model. We name SDGL without different components as follows:
\begin{itemize}
	\item SDGLw/oGLoss: SDGL without Grpah regularization. We set the factors $\lambda$ directly to 0.
	\item SDGLw/oDyAdj: SDGL without Dynamic graph learning layer.
	\item SDGLw/o IFM: SDGL without the Information fusion module in Dynamic Graph Learning Layer. We pass the node-level inputs to the dynamic graph learning module, which means consider only dynamic node inputs. 
	\item SDGLw/o IFM +: We replace the information fusion module with a summation operation.
\end{itemize}

The test results measured using RMSE, MAPE are shown in Figure \ref{fig4}(a), more details are detailed in Appendix A.6. Several observations from these results are worth highlighting:
\begin{itemize}
	\item The best result on each dataset is obtained with SDGL, proving that each of our modules worked.
	\item Removing the dynamic graph learning layer (DGLL) causes performance drop, which is obvious in the RMSE for PEMSD4 dataset and MAPE for PEMSD8 dataset. Also after removing the DGLL, the effect gradually deteriorates over time among the 12 prediction steps, as shown in Figure \ref{fig4}(b). We conjecture the reason is that long-term prediction lacks sufficiently useful information from historical observations and thus benefits from the short-term patterns learned by DGLL to deduce future values. 
	\item After removing the IFM module, performance drops when dynamic graph are constructed base on only node-level inputs, especially in the long term (e.g., 60 Min) prediction. Whereas with the + operation instead of information fusion, the performance drops only slightly in all prediction steps. It demonstrates the importance of long-term patterns have significant impacts on inter-variate dependencies.
	\item The performance drops after removing graph regularization, proving that the regularization improves the quality of learned graph by SDGL. But overall, SDGLw/oGLoss outperformed AGCNR in multi-step prediction on both datasets, except for MAPE in 60-min prediction. It suggests that our proposed static- and dynamic- graph approach itself can improve performance, rather than relying on the addition of graph regularity.
\end{itemize}
\begin{figure}[t]
	\centering
	\includegraphics[width=0.9\columnwidth]{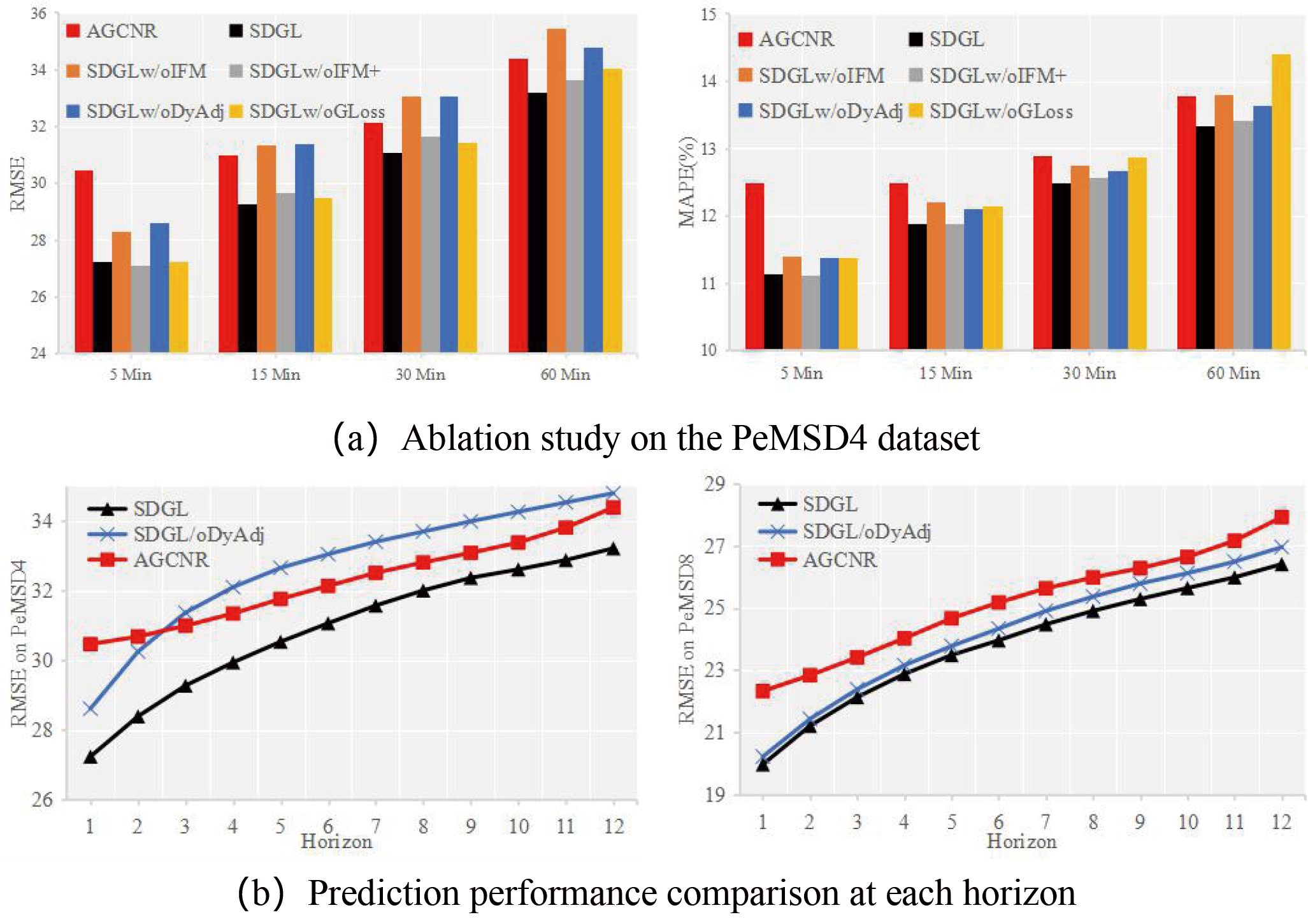}\\ 
	\caption{Results of SDGL in the ablation tests}
	\label{fig4}
\end{figure}

\subsection{Study of the Graph Learning Layer}
\begin{figure}[t]
	\centering
	\includegraphics[width=0.9\columnwidth]{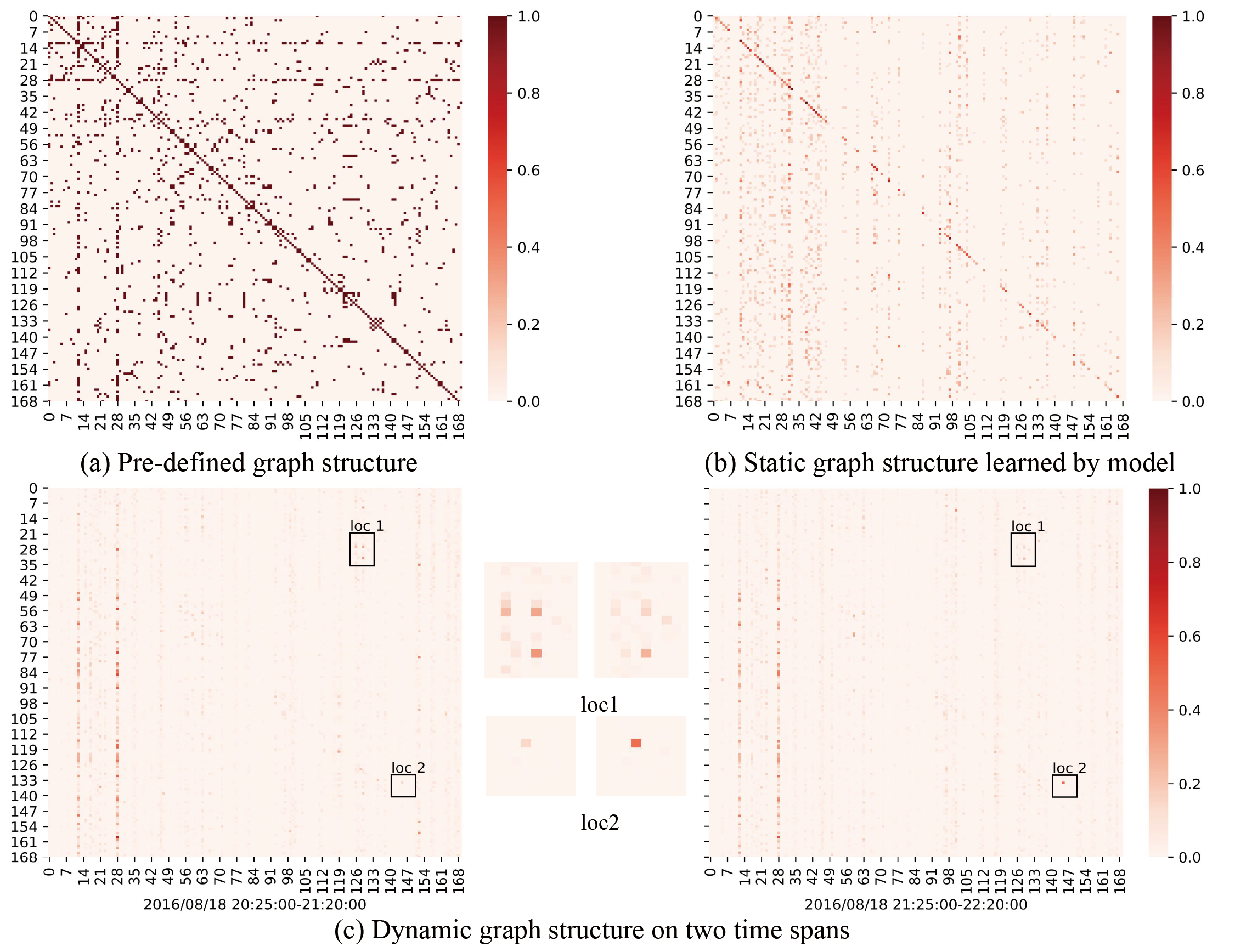}\\ 
	\caption{Graph structure visualization}
	\label{fig5}
\end{figure}

To verify the effectiveness of our proposed graph learning method, we visualize and analyze the matrices learned by model. Figure \ref{fig5} shows the pre-defined graph matrix in PeMSD8 dataset, static graph structure and the dynamic graphs at two different time spans. As shown in Figure \ref{fig5}(b), we observe that in static graph, most of nodes present self-attention, i.e. diagonal line in the figure. In contrast to the manual addition of self-loops in the predefined matrix, self-attention in static graph is acquired by model. Further, we present the dynamic graph matrices at different time spans in Figure \ref{fig5}(c). We find that there are two data points of higher weight in the dynamic graph, which corresponds to exactly two important data points in the predefined graph, i.e., 12 and 28. As shown in Figure \ref{fig5}(c), the heatmaps of the dynamic matrices on the two close time spans are very close to each other, which indicates that the dynamic graphs learn a short-term pattern. The two regions marked with “loc 1” and “loc 2” demonstrate that our model can capture the changes in intervariate dependencies.
\section{Conclusion}
In this paper, we propose a novel graph neural network for multivariate time series forecasting. This work models long- and short-term spatiotemporal patterns in multivariate time series data by constructing static and dynamic graph matrices from data. Also dynamic graphs are generated dynamically based on time-varying inputs and fixed long-term patterns to capture the changing spatial-temporal dependence among variables. Our approach shows excellent performance in both multi-step traffic prediction and single-step time series prediction tasks. It reveals that correctly and adequately modeling the dependencies between pairs of variables is essential for understanding time series data.

\bibliography{aaai22}
\end{document}